\documentclass[runningheads,a4paper]{llncs}
\usepackage{tablefootnote}
\usepackage{amssymb}
\usepackage{graphicx}
\usepackage{comment}
\usepackage{multirow}
\usepackage{subfig}
\usepackage[sorting=none]{biblatex}
\usepackage{amsmath}
\usepackage[bottom]{footmisc}
\usepackage{array}
\newcommand{\PreserveBackslash}[1]{\let\temp=\\#1\let\\=\temp}
\newcolumntype{C}[1]{>{\PreserveBackslash\centering}p{#1}}
\newcolumntype{R}[1]{>{\PreserveBackslash\raggedleft}p{#1}}
\newcolumntype{L}[1]{>{\PreserveBackslash\raggedright}p{#1}}

\usepackage{url}
\newcommand{\keywords}[1]{\par\addvspace\baselineskip
\noindent\keywordname\enspace\ignorespaces#1}

\begin{document}

\mainmatter  

\title{Densely-Populated Traffic Detection using YOLOv5 and Non-Maximum Suppression Ensembling}
\titlerunning{Densely Populated Traffic Detection using YOLOv5 and NMS Ensembling}
%
%
%
\author{Raian Rahman \and Zadid Bin Azad \and
Md. Bakhtiar Hasan}
\authorrunning{Rahman et al.} 
%
\tocauthor{Raian Rahman, Zadid Bin Azad and Md. Bakhtiar Hasan}
\institute{Department of Computer Science and Engineering\\
Islamic University of Technology\\
\email{\{raianrahman, zadidbinazad, bakhtiarhasan\}@iut-dhaka.edu}}

%


%
%

\maketitle

\begin{abstract}

Vehicular object detection is the heart of any intelligent traffic system. It is essential for urban traffic management. R-CNN, Fast R-CNN, Faster R-CNN and YOLO were some of the earlier state-of-the-art models. Region based CNN methods have the problem of higher inference time which makes it unrealistic to use the model in real-time. YOLO on the other hand struggles to detect small objects that appear in groups. In this paper, we propose a method that can locate and classify vehicular objects from a given densely crowded image using YOLOv5. The shortcoming of YOLO was solved my ensembling 4 different models. Our proposed model performs well on images taken from both top view and side view of the street in both day and night. The performance of our proposed model was measured on Dhaka AI dataset which contains densely crowded vehicular images. Our experiment shows that our model achieved mAP@0.5 of $0.458$ with inference time of $0.75$ sec which outperforms other state-of-the-art models on performance. Hence, the model can be implemented in the street for real-time traffic detection which can be used for traffic control and data collection.


\end{abstract}
\footnotetext{Accepted for Springer Lecture Notes on Data Engineering and Communications Technologies as a part of BIM 2021}

\section{Introduction}
An increasing number of vehicle types in urban areas pose many problems like traffic congestion, long queue in toll and parking sites. To solve traffic problems in mega-cities and to monetize traffics in areas like toll booths, parking lots, and analyzing types of vehicles in a city more efficiently and effectively, an intelligent system is required. As an indispensable part of the intelligent traffic monitoring system, accurate vehicle detection and real-time performance is the most challenging part which is gaining the attention of researchers all over the world. Efficient vehicle detection and classification in densely populated areas can facilitate automated toll collection, smart parking systems, and identification of vehicles related to crimes.

The task of vehicle detection can be formulated as a multi-object detection problem. In simple terms, object detection is the task of locating the objects in an image with a bounding box and detecting the class of that object. For this, convolutional neural network (CNN) based methods have been widely used in the recent past. The prominent state-of-the-art methods utilize R-CNN \cite{girshick2014rich}, Fast R-CNN \cite{10.1109/ICCV.2015.169}, Faster R-CNN \cite{7485869} to achieve this task. But the problem with these two-stage-based models is that training happens in multiple phases and the network is too slow at inference time, which impedes real-time detection of vehicles. To solve this problem, recently You Only Look Once (YOLO) \cite{7780460} introduced a faster way of real-time object detection making it usable in real-life applications. However, this architecture struggles to detect small objects that appear in groups \cite{7780460}. 

To solve this issue, we trained 4 separate models that utilize the ensemble technique to aggregate the separate predictions using Non-Maximum Suppression. Our contributions are as follows:
\begin{itemize}
    \item Trained a total of 4 YOLOv5 \cite{glenn_jocher_2021_4679653} models using different image sizes and hyper-parameters.
    \item Aggregated the prediction of 4 models using ensemble model that facilitates faster detection of vehicles.
    \item Introduced additional difficulty by adding low-light nighttime images and top-view images with densely crowded vehicles to training samples to improve the accuracy and robustness of the model.
\end{itemize}
These steps resulted in a solution that can be used in real-time and low light situations even in densely populated streets. Besides it also ensured that our solution outputs a result with acceptable accuracy which makes our model usable in congested and complex scenes.

\section{Related Work}
The traditional approaches\cite{laopracha2017comparative, cao2011linear} for vehicle detection apply common machine learning algorithms like the histogram of oriented gradient (HOG) to extract features from vehicle images. After extracting the features, the vehicles are then classified using Support Vector Machine (SVM). Other approaches use Deformable Part Model (DPM)\cite{pan2016study} to detect vehicles. Even though these approaches provide comparable accuracy, they involve handcrafted feature designing that requires human intervention.

Recent advances in deep learning facilitated by the availability of large datasets and big compute have made them a viable option for vehicle detection. Earlier approaches \cite{tang2017vehicle, gao2017scale, huttunen2016car} utilize Convolutional Neural Network (CNN) to perform feature extraction and softmax function for classification. Later, more efficient models like R-CNN \cite{girshick2014rich} and fast R-CNN\cite{10.1109/ICCV.2015.169} and Faster R-CNN\cite{7485869} models were proposed. All these models utilize a Region-based Convolutional Neural Network, which uses a technique called Selective Search \cite{uijlings2013selective} to select a small number of candidate regions among all possible regions. As a result, the model requires running an image classification algorithm for a smaller amount of region making the model run faster. R-CNN is comparatively slower among all three models as it generates lots of candidate regions. Fast R-CNN \cite{10.1109/ICCV.2015.169} addressed this issue by feeding the input image to a CNN to generate a convolutional feature map. Then the candidate regions are proposed using an RoI pooling layer and feeding it into a fully connected network. The number of candidate regions proposed by Fast R-CNN is less than that of R-CNN, hence it requires less time for inference. But the Selective Search algorithm, used by Fast R-CNN, is not a machine learning algorithm, so it cannot learn from the context, and often proposes a bad candidate for the region. Later, Faster R-CNN \cite{7485869} was proposed with the idea of replacing selective search as it is a time-consuming process. Faster R-CNN provides the fastest running time compared with R-CNN and Fast R-CNN.  However, it is still not fast enough to detect objects in real-time. Additionally, all these three models require huge computation due to having a complex model containing a large number of parameters. 

Recently, YOLO is being used for vehicle detection \cite{a14040114,sang2018improved,asha2018vehicle}. Instead of using the region selection method, YOLO uses Convolutional Neural Network that predicts the bounding boxes as well as the class for these boxes. It divides the image into an $S \times S$ grid where $S$ is a constant value. For each grid, YOLO generates a constant number of bounding boxes. Then if a bounding box has confidence greater than a certain threshold, the bounding box is selected to locate the object within the image. YOLO is by far the fastest algorithm for vehicle detection and its speed is helpful to implement real-time vehicle detection systems.

\section{Proposed Methodology}

\begin{figure}[t!]
     \centering
     \includegraphics[width = \textwidth]{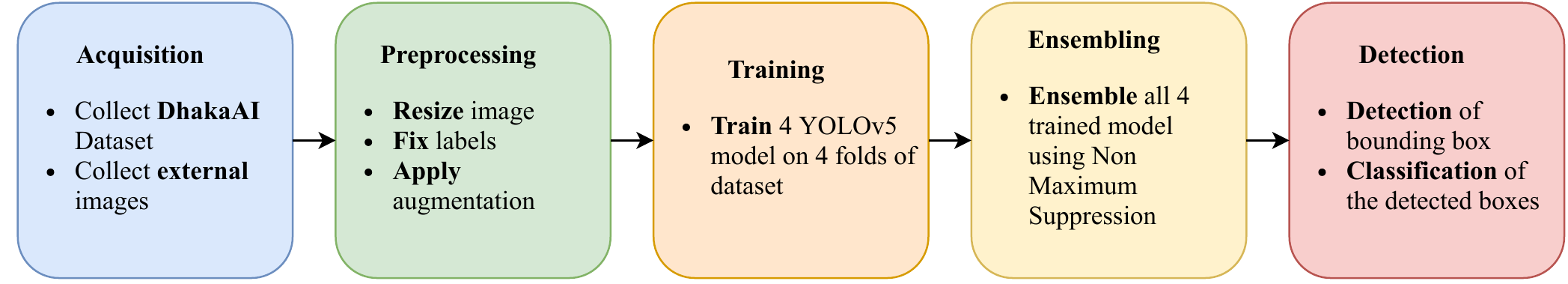}
     \caption{The pipeline of our proposed solution: First, we acquired the training data. During preprocessing, images were resized, relabelled before creating four different folds of the dataset. Different augmentation technique was applied to these folds. During training, these folds were trained independently with YOLOv5 model with different setups. All four of our trained models were then ensembled using Non Maximum Suppression. The last state of our work is the images with bounding box surrounding the vehicle objects of an image with its class}
     \label{steady_state}
\end{figure}

\subsection{Overview}
Our proposed method consists of 3 main modules. First, we acquired and preprocessed the dataset. During preprocessing, we applied augmentation, resized the images into uniform shapes, and created training and testing folds. Then, four different models were trained with these different training folds. After training, we ensembled the models using Non-Maximum Suppression \cite{1699659} for final inference. A complete pipeline of our proposed methodology is illustrated in \figurename~\ref{steady_state}.

\subsection{Dataset Acquisition}
For this experiment, we used the ``DhakaAI'' \cite{DVN/POREXF_2020} dataset developed under the ``Dhaka AI 2020 challenge''. The dataset consists of 3000 annotated images of traffic objects. The training dataset consisted of 21 classes. The most challenging part about the dataset is it contains images of vehicles from a different point of view. There were images from the front view, back view, side view and most importantly top view of streets. We also added around 200 new images for training to increase the sample of rare class vehicles. These new images were hand-annotated using labelImg tool \cite{Tzutalin.LabelImg}. Most of these images were top-view nighttime images.

\subsection{Preprocessing}
For generalizing a model for object detection using deep learning architecture, a prerequisite is to have enough training examples for each class so that the model can learn properly. But, after exploring the DhakaAI dataset \cite{DVN/POREXF_2020}, we found that it has a huge class imbalance. The number of labels for each class is shown in \tablename~\ref{tab:class-label}. Here, some of the classes have less than 50 samples in the training dataset.

\begin{table}[b!]
    \centering
    \caption{Sample Distribution per Class}
    \label{tab:class-label}
    \begin{tabular}{L{3.5cm} C{1.2cm} | L{3.5cm} C{1.2cm}}
        \hline
        \textbf{Class Name} & \textbf{Label Count} & \textbf{Class Name} & \textbf{Label Count}\\\hline
        Ambulance & 76 & Pickup & 1178\\
        Army Vehicle & 25 & Police Car & 33\\
        Auto Rickshaw & 465 & Rickshaw & 3495\\
        Bicycle & 465 & Scooter & 30\\
        Bus & 3340 & SUV & 667\\
        Car & 5574 & Taxi & 59\\
        Garbage Van & 8 & Three Wheeler (CNG) & 2982\\
        Human Hauler & 170 & Truck & 1475\\
        Minibus & 100 & Van & 682\\
        Minivan & 815 & Wheelbarrow & 251\\
        Motorbike & 2252 &  & \\
        \hline
    \end{tabular}
\end{table}

To resolve this issue, we used augmentation using tools from Roboflow \footnote{Available at: \url{https://roboflow.com/}} and Albumentations library \cite{info11020125} for image augmentation. Although augmentation did not provide a very good result in the case of densely populated images, it improved the result in the case of night images. 

\begin{figure}[b]
     \centering
     \subfloat[][]{\includegraphics[width = 1.5 in]{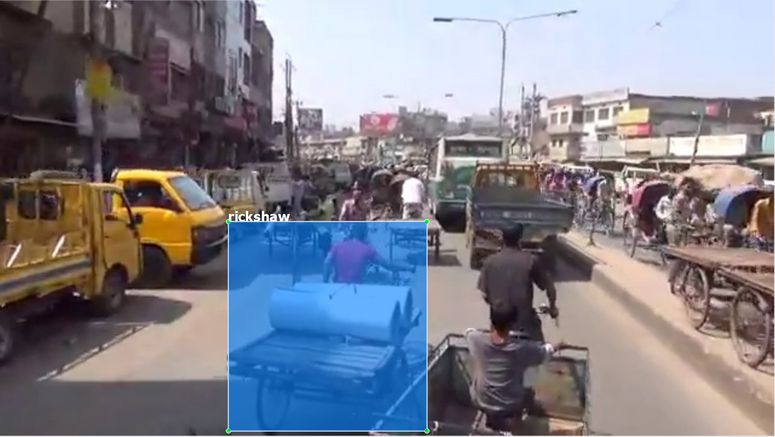}\label{<figure100>}}
     \hspace{.05 in}
     \subfloat[][]{\includegraphics[width = 1.5 in]{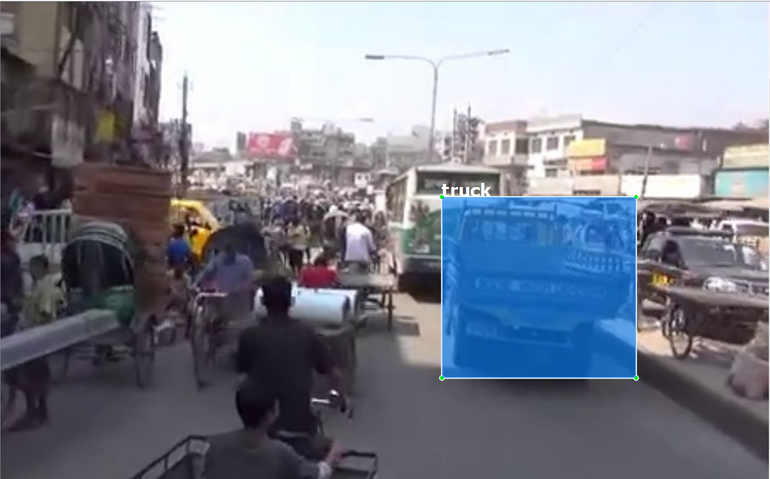}\label{<figure101>}}
     \hspace{.05 in}
     \subfloat[][]{\includegraphics[width = 1.5 in]{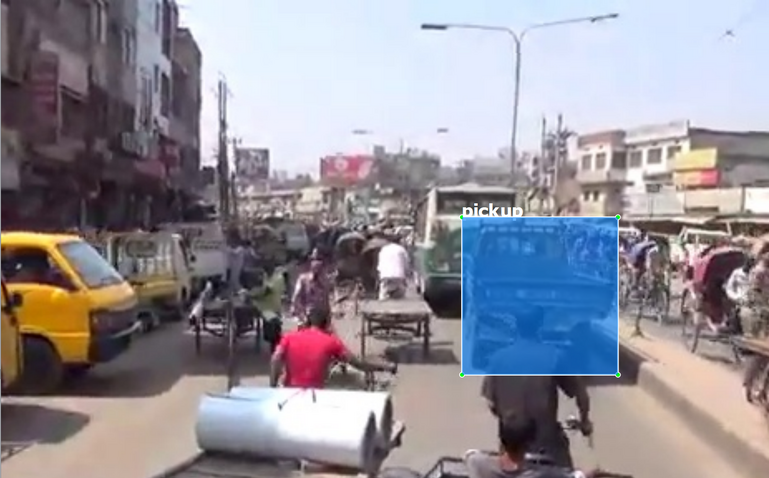}\label{<figure102>}}
     \caption{The figure shows an anomaly in the DhakaAI dataset. In (a), we can see that a wheelbarrow is labeled as a rickshaw. In (b) and (c), we can see that the same vehicle, which is a pickup is labeled as a truck and on the next image it is labeled as pickup.}
     \label{mislabeled}
\end{figure}

During the exploration of the dataset, we found that there was a lot of mislabelling in the DhakaAI dataset training data. We also found that two images had different labeling of class for the same car in the same frame (illustrated in \figurename~\ref{mislabeled}). So, we hand-annotated all 3000 images and labeled all the mislabelled objects as well as fixed labeling of wrongly labeled objects in the image.

\begin{table}[t]
    \centering
    \caption{Image resolution and applied augmentation for different folds of training dataset. All images had $1024 \times 1024$ resolution.}
    \label{tab:dataset-split}
    \begin{tabular}{C{1cm} C{2.5cm} C{2.5cm} C{2.5cm}}
        \hline
        \textbf{Fold No.} & \textbf{Train Set Image Count} & \textbf{Validation Set Image Count} & \textbf{Augmentation}\\
        \hline
        1 & $2506$ & $600$ & Sharpened\\
        2 & $2321$ & $785$ & Sharpened\\
        3 & $2400$ & $706$ & Sharpened\\
        4 & $1200$ & $400$ & Darkened and Sharpened\\\hline
    \end{tabular}
\end{table}

Another challenge in the dataset is it does not have uniform image quality. Some of the images are in landscape mode while some of the images are in portrait mode. So, we resized the images to $1024 \times 1024$ pixels.

For the train and validation set split, we used the k-fold Cross-Validation technique so that our model could learn from the complete dataset. While creating the fold, we tried to make sure that images from the same frame in the train split do not occur in the validation split. Count of train-validation split for each fold is given in \tablename~\ref{tab:dataset-split}.

\subsection{Model Selection}

Although the key priority of our work was to localize and classify the vehicular objects on a street image, we also had to look into the inference speed so that it could be implemented real time. We had to discard R-CNN, Fast R-CNN and Faster R-CNN as it could not compete with YOLO models in both performance and inference time. YOLO on the other hand, YOLO had a much less inference time with better accuracy. Among different versions of YOLO we chose YOLOv5\cite{glenn_jocher_2021_4679653} due to its simple architecture compared to R-CNN based models. Even YOLOv5 is faster and more robust than other members of YOLO family. 


Even if the author of YOLOv4\cite{bochkovskiy2020yolov4} got official approval of YOLO, the version of YOLOv5 \cite{glenn_jocher_2021_4679653} developed by Ultralytics LLC team did not get any acknowledgement from the original author of YOLO. Still YOLOv5 provides much better performance compared to other models of YOLO family\cite{liu2020research}. YOLOv5 inherits the advantages of YOLOv4 \cite{bochkovskiy2020yolov4} by adding SPP-NET \cite{7005506} along with some enhancement techniques. YOLOv5 has become the new state of the art for object detection\cite{yan2021real}. YOLOv5 was mainly developed to balance between real-time performance and detection accuracy. 

YOLOv5s \cite{glenn_jocher_2021_4679653} , YOLOv5m \cite{glenn_jocher_2021_4679653}, YOLOv5l \cite{glenn_jocher_2021_4679653}, YOLOv5x \cite{glenn_jocher_2021_4679653} are the four versions of YOLOv5 where YOLOv5s being the lightest model and YOLOv5x being the heaviest model respectively. Among all these four version, there is a trade off between the detection speed and real-time performance.  The key difference among these versions are the number of feature extraction modules and convolution kernel in specific location of the network. 

The network consists of three networks. These are: backbone network, neck network and detect network. Backbone network is a convolutional neural network for aggregating the fine-grained images and forming image features. Neck network is responsible for combining the image features collected by backbone network and transmitting the feature map to the detect network. The detect network is responsible for detection and classification part of the model. It applies anchor boxes on the feature map from the neck network. It also contains a softmax layer which predicts the probability of the class the bounding box surrounding the object.

For image enhancement, YOLOv5 uses mosaic data augmentation to solve the low dataset problem. It applies operations like random inversion, zooming, cropping on four images and then combines them into a single image.

In traffic detection, the core priority was to improve the performance, so we chose YOLOv5x for our training model. It contains $607$ layers with $88,568,234$ trainable parameters. The model was pre-trained using Common Object in Context (COCO) dataset \cite{lin2014microsoft} to detect $80$ classes. For our task, we changed the final layer to detect only $21$ classes corresponding to the $21$ vehicle classes available in the DhakaAI dataset.

\subsection{Ensemble Learning}
To ensure the robustness and accuracy of our model, we trained 4 separate models using different sets of images. Each of the models proposes multiple bounding boxes to specify candidate regions for vehicle detection. We used Non-Maximum Suppression \cite{1699659} to aggregate these bounding boxes to select the ones having the most confidence. The way it works is the system takes all the bounding boxes proposed by all four models and puts them in a priority queue sorted based on the confidence of the models predicting them. It then selects the box with the highest confidence from the queue and calculates the Intersection over Union (IoU) with the rest of the boxes. If the IoU value exceeds a certain threshold for any of the remaining boxes, that box is discarded. It then removes the bounding box with the highest confidence from the queue and adds it to the selected box list. This process is repeated until there is no bounding box remaining in the priority queue. Finally, the boxes in the selected box list are returned.
\section{Result and Analysis}

\subsection{Experimental Setup}

During training we had to train four different models and ensembled these four models for final output. All four of our model was trained on Google Colab\cite{Bisong2019}. Google Colab provides cloud based training utility with free GPU access for limited amount of time. For each fold of our dataset, we trained a model. The first three models were trained with image resolution of $1024 \times 1024$ pixels. But the fourth model was trained with image resolution of $640 \times 640$ pixels. First three fold of our dataset contained all the images while the fourth fold contained only the night images. On the dataset, it was seen that the night images itself were quite distorted noisy. So. we decided to train the night images on lower resolution as it might then focus on the  the larger objects of the night images. Also thus we could train our model for longer time. 

All four of our models were trained with Tesla T4 GPU which comes with $16$ GB of video memory. All four of our model was trained for around $12$ hours each. 

For training, we used a YOLOv5 implementation by the Ultralytics \footnote{Available at: \url{https://github.com/ultralytics/yolov5}}. We used Stochastic Gradient Descent as our optimizer. Image augmentation parameters used for each of the models are given in \tablename~\ref{tab:hyp_tables}.

\begin{table}[t]
    \centering
    \caption{Training Specification for Each Model. All 4 models had Stochastic Gradient Descent optimizer with a learning rate $0.01$ and momentum $0.937$}
    \label{tab:training_specification}
    \begin{tabular}{C{1cm} C{1.8cm} C{2cm} C{2.7cm} C{1.2cm}}
    \hline
    \textbf{Model} & \textbf{Training Data} & \textbf{Image Size} & \textbf{Number of Epochs} & \textbf{Batch Size}\\\hline
    1 & Fold 1 & $1024 \times 1024$ & 80 & 4 \\
    2 & Fold 2 & $1024 \times 1024$ & 80 & 4\\
    3 & Fold 3 & $1024 \times 1024$ & 80 & 4\\
    4 & Fold 4 & $640 \times 640$ & 120 & 16\\\hline
    \end{tabular}
\end{table}

\begin{table}[t]
    \centering
    \caption{Image augmentation parameters during training.}
    \label{tab:hyp_tables}
    \begin{tabular}{l c}
        \hline
         \textbf{Hyperparameter} & \textbf{Value}\\\hline
         Image HSV - Hue augmentation & $0.015$\\
         Image HSV - Saturation augmentation & $0.7$\\
         Image HSV - Value augmentation & $0.4$\\
         Image Rotation & $5.0$\\
         Image Translation & $0.1$\\
         Image Scale & $0.5$\\
         Image Flip Left-Right - Probability & $0.5$\\
         Image Mosaic - Probability & $1.0$\\
         Image Mixup - Probability & $0.0$\\\hline
    \end{tabular}
\end{table}

\subsection{Evaluation Metrics}

To evaluate our performance, we used mean Average Precision (mAP) over training epochs following . The formula for calculating mean average precision for object detection is
\begin{align}
    maP = \frac{1}{n}\sum_{k=1}^{k=n} AP_k
\end{align}
where $n$ is the number of classes and $AP_k$ is the average precision for class $k$. Average precision (AP) is a way of summarizing the precision-recall curve into a single value representing average of all precision. The formula for calculating AP is 
\begin{align}
    AP@n = \sum_{k=0}^{k=n-1}[Recall(k)-Recall(k+1)]\times Precision(k)
\end{align}
where $n$ is the number of thresholds and $Recall(n)=0$ and $Precision(n)=1$. We used the checkpoint where the model had most \emph{mAP@0.5}

     

For inference, we ensembled the weight of all four models we trained. We used Non-Maximum Suppression during the ensemble and the confidence threshold for each predicted bounding box was set to $0.3$


\subsection{Result Discussion}

We used all four training model's weights during the final inference. We ran an inference on test data - 2 provided by DhakaAI. We hand-annotated $450$ test images and executed inference. On that test, our model achieved  $mAP@0.5$ value of $0.458$. We also conducted inference on one of our validation sets. During validation set inference, our model achieved $mAP@0.5$ value of $0.883$ and $mAP@0.95$ value of $0.677$.

We compared the result of our model with other models of YOLO family as well as Faster R-CNN model. \tablename~\ref{tab:comparison-result} shows the comparison between these models. For comparison we compared our model's performance as well as the inference time on a single image with YOLOv3, YOLOv4 and Faster R-CNN. We trained each of these models for $12$ hours on google colab in a similar environment. The table shows that, our model has achieved the most $mAP@0.5$. As our proposed method ensembles $4$ different models during inference, the inference time of our solution is a little bit higher compared to other models. Still the precision performance of our model outperforms the other models.

\begin{figure}[b!]
     \centering
     
     \subfloat[][]{\includegraphics[width = 1.4 in]{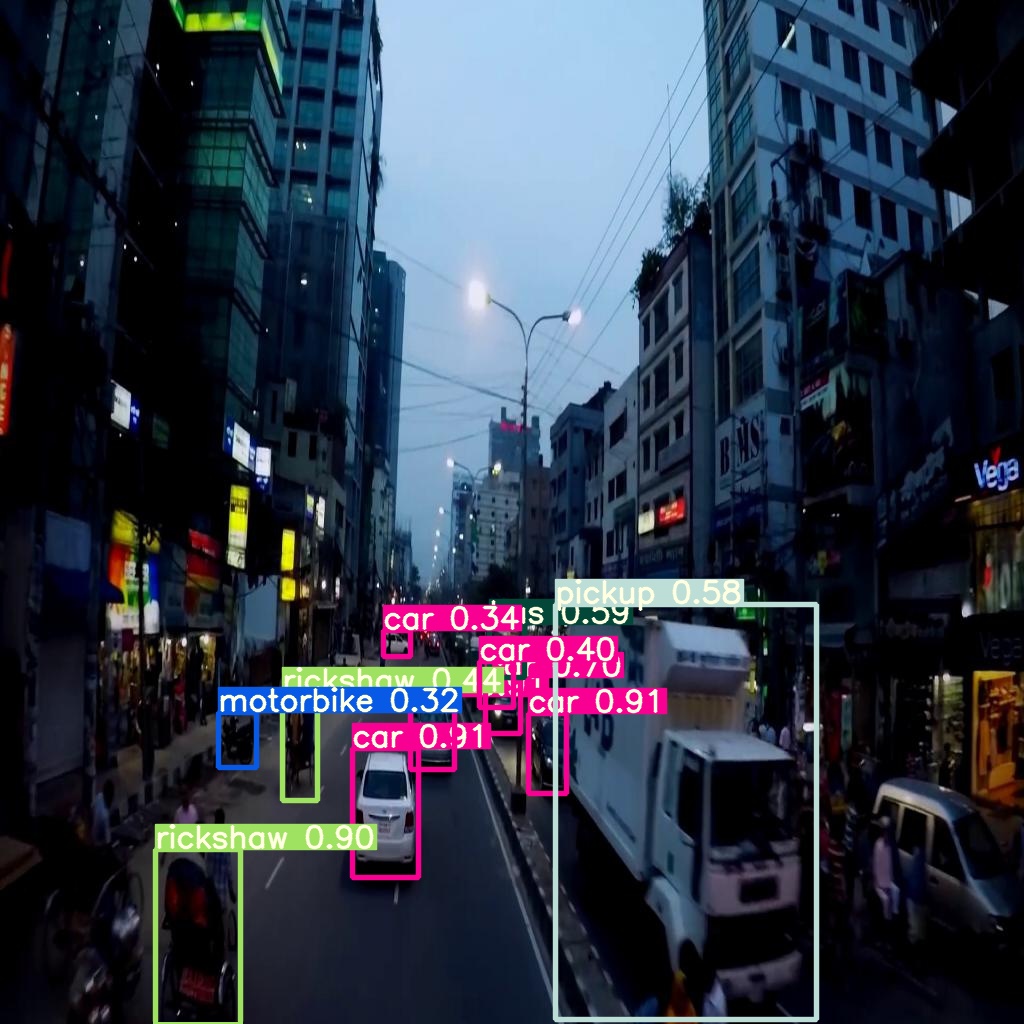}\label{<night1>}}
     \hspace{.1 in}
     \subfloat[][]{\includegraphics[width = 1.4 in]{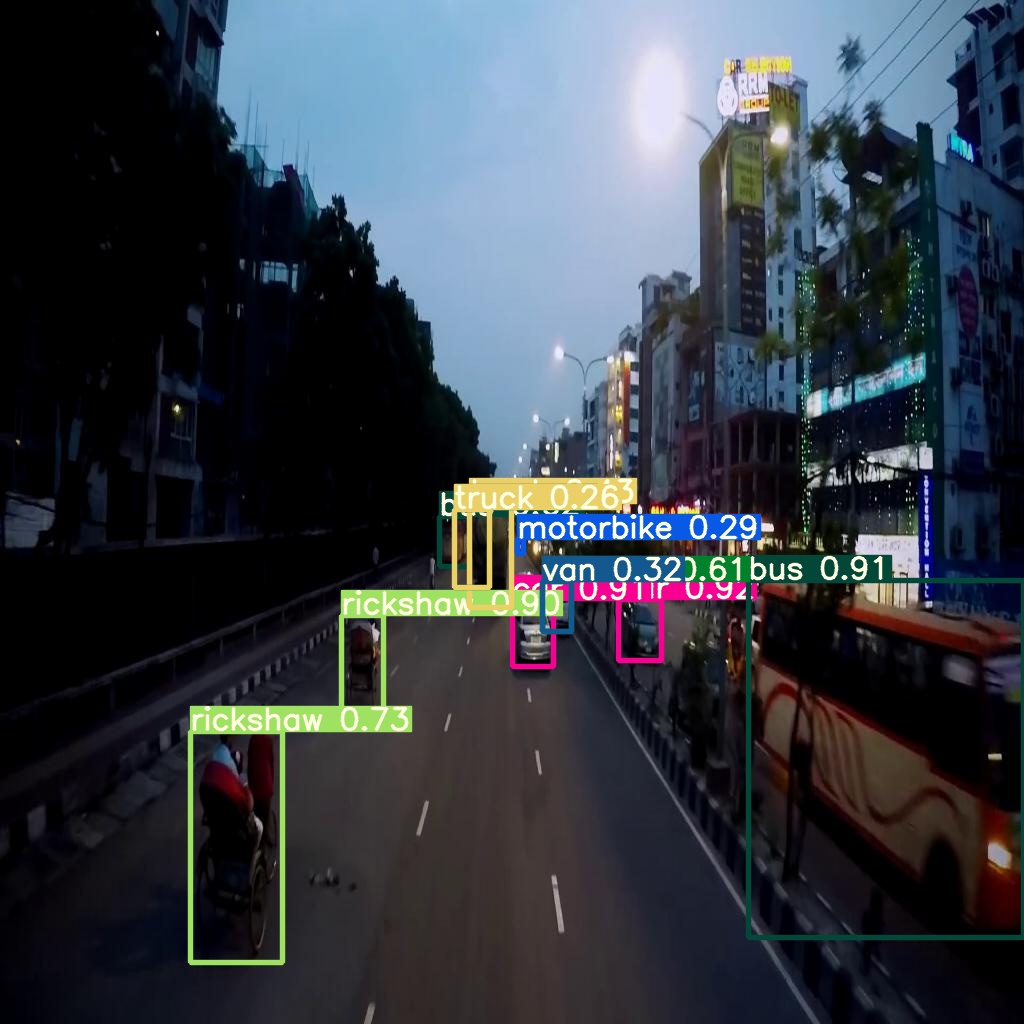}\label{<night2>}}
     \hspace{.1 in}
     \subfloat[][]{\includegraphics[width = 1.4 in]{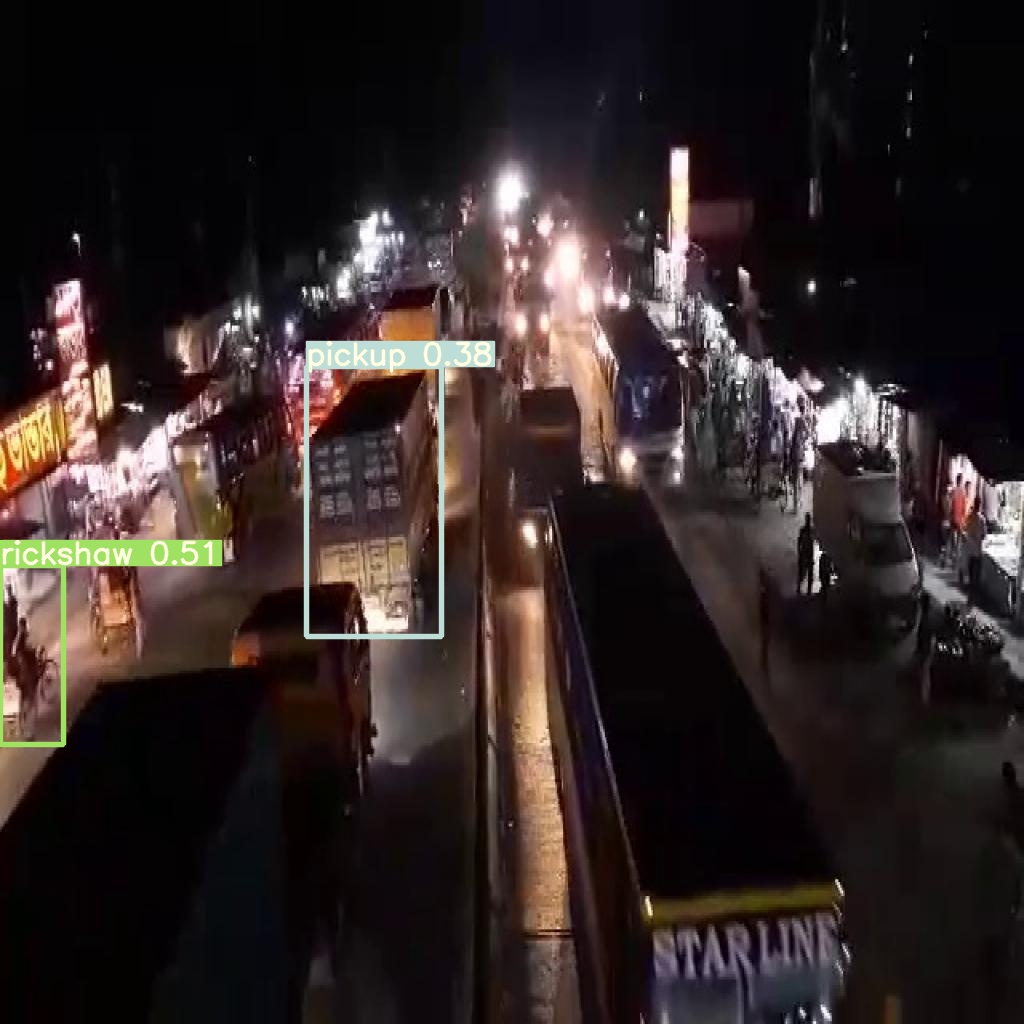}\label{<night3>}}
     \caption{Performance on night image in densely populated image \figurename~\ref{<night1>} and less densely populated image \figurename~\ref{<night2>}. In \figurename~\ref{<night3>}, it can be seen that the model performs poorly in very low light sample}
     \label{night-image}
\end{figure}
\begin{figure}[t]
     \centering
     \subfloat[][]{\includegraphics[width = 1.4 in]{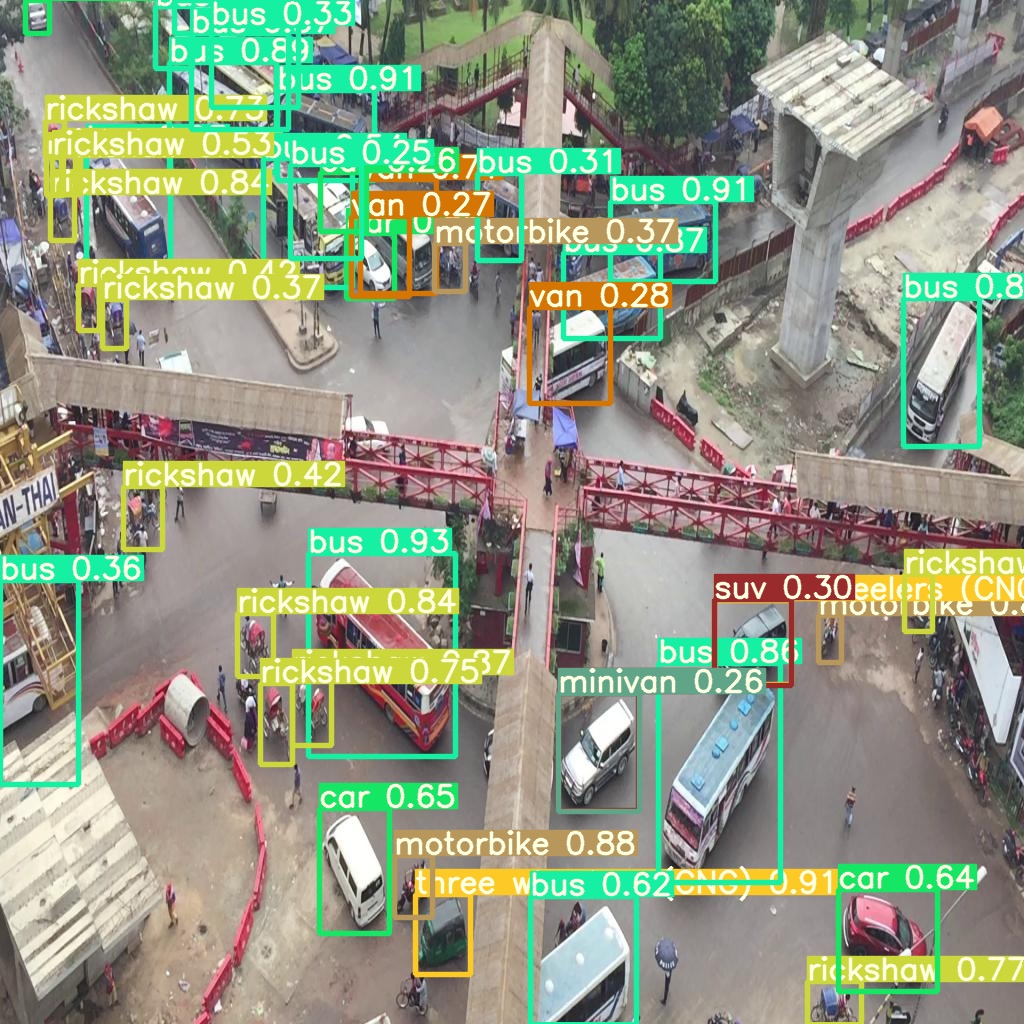}\label{<figure4>}}
     \hspace{.1 in}
     \subfloat[][]{\includegraphics[width = 1.4 in]{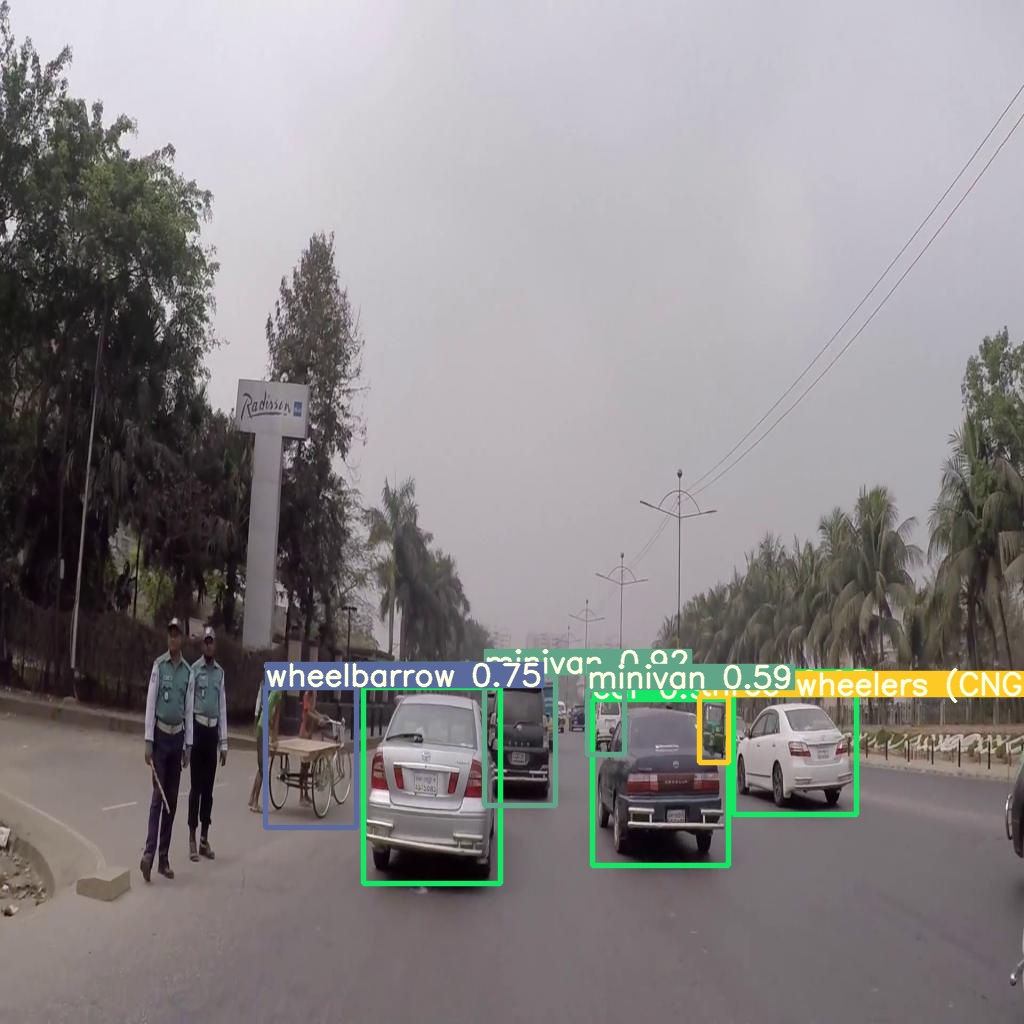}\label{<figure5>}}
     \hspace{.1 in}
     \subfloat[][]{\includegraphics[width = 1.4 in]{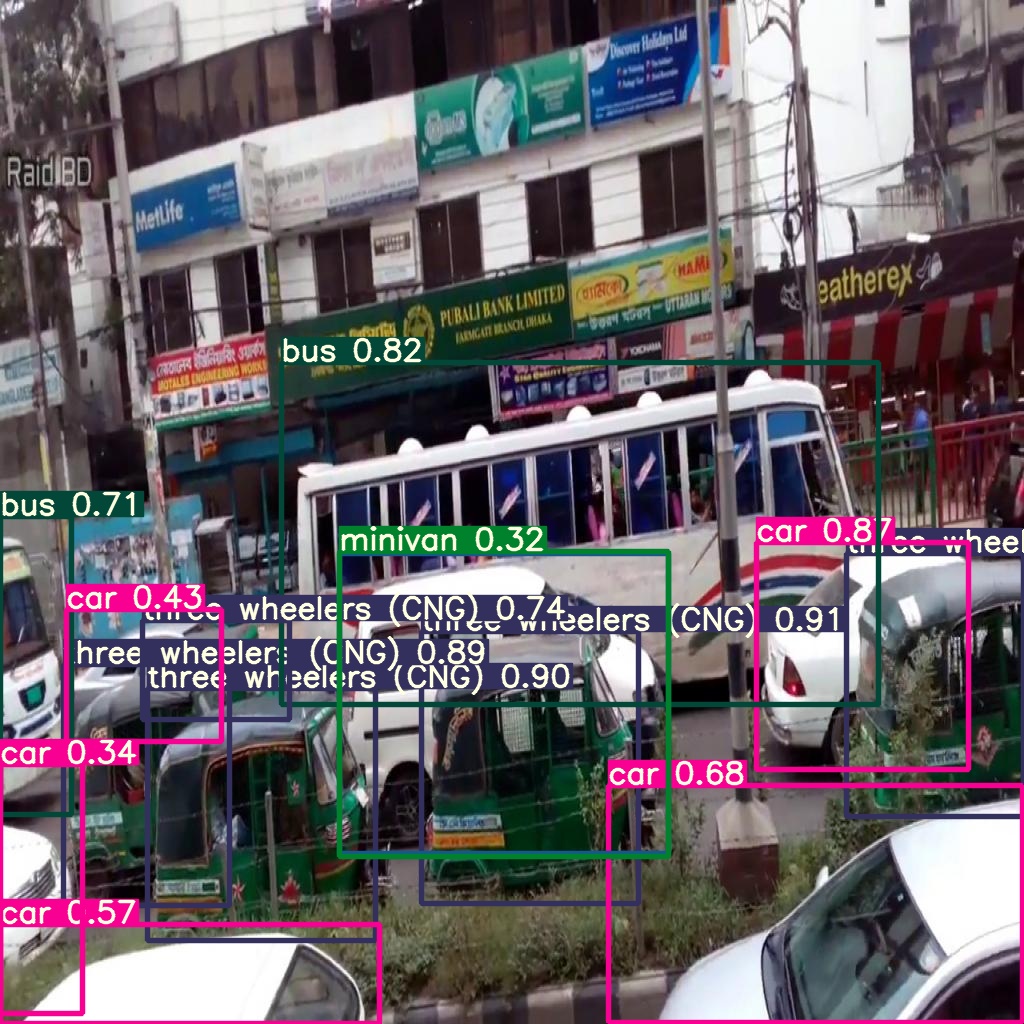}\label{<figure6>}}
     
     \caption{Performance on images taken from different views of street.}
     \label{top-view}
\end{figure}

\begin{table}[t]
    \centering
    \caption{Comparison of performance and inference time with other models. Here Faster R-CNN. YOLOv3. YOLOv4 and YOLOv5x show performance trained on a single fold of train dataset while YOLOv5x with NMS ensembling model shows the result of our four combined models.}
    \label{tab:comparison-result}
    \begin{tabular}{l c C{2cm}}
    \hline
    \textbf{Model Name} & \textbf{mAP@0.5} & \textbf{Inference Time (s)}\\
    \hline
    Faster R-CNN & 0.356 & 0.39\\
    YOLOv3 & 0.266 & 0.18\\
    YOLOv4 & 0.313 & 0.28\\
    YOLOv5x & 0.372 & 0.14\\
    YOLOv5 with NMS ensembling (ours) & 0.458 & 0.75\\\hline
    \end{tabular}
\end{table}
Output of our model for different scenario is illustrated in \figurename~\ref{night-image} and  \ref{top-view}.
Our model was able to localize and detect most of the vehicular objects for a given image taken from a different view of the street. It also performed well in the case of night images. \figurename~\ref{night-image} illustrates the performance of our model on night images. It could locate most of the vehicles as well as properly classify those vehicles in both densely populated images and in less populated images. However, as seen in \figurename~\ref{<night3>}, our model could not detect most of the vehicles in a very low-light noisy image.

In \figurename~\ref{top-view}, we illustrated our model's performance on images taken from a different view of the streets which shows it can locate and detect the objects properly. As seen in the figure, the model also performs well in the case of occluded objects in the image. 

And our model could run inference within $0.75$ second per image. So, it could also be implemented in real-time vehicular traffic detection applications. 


\section{Conclusion}

This paper proposed a new method of traffic object detection using YOLOv5. To improve the performance and robustness of our method, we ensembled 4 different models using Non-Maximum Suppression ensembling. We also tried to incorporate dataset modification by adding night images from different view-angles. Our experiment compared the performance of our model with other state of the art models on Dhaka AI dataset. Result shows that our model had better precision. Due to limited resources, we couldn't test our model's performance on other baseline dataset. For further experimentation, our work could be expanded on how we can use better ensembling methods like weighted ensembling or voting mechanism for faster inference time.



\section{Acknowledgement}
We would like to thank Redwan Karim Sony, Department of Computer Science and Engineering, Islamic University of Technology and Mohammad Sabik Irbaz, Pioneer Alpha Limited for their continuous support and suggestions throughout the work. We would also like to thank the organizing committee of Dhaka AI 2020 for organizing the competition.

\medskip
\printbibliography
\end{document}